\documentclass[10pt,letterpaper]{article}

\usepackage{cogsci}

\cogscifinalcopy 

\usepackage{graphicx}
\usepackage{amsmath}
\usepackage{pslatex}
\usepackage{tikz}
\usepackage{subcaption}

\usepackage{color,soul}
\usepackage{microtype}

\usepackage[hidelinks]{hyperref}
\usepackage{apacite-mod}
\usepackage{float} 

\newcommand{\Getal}{Gwilliams et al.}

\title{A predictive learning model can simulate temporal dynamics and context effects found in neural representations of continuous speech}
 
\author{
{\large \bf Oli Danyi Liu\textsuperscript{1}, Hao Tang\textsuperscript{1}, Naomi H. Feldman\textsuperscript{2}, Sharon Goldwater\textsuperscript{1}}
\\ oli.liu@ed.ac.uk, hao.tang@ed.ac.uk, nhf@umd.edu, sgwater@inf.ed.ac.uk
\\ \textsuperscript{1}University of Edinburgh  \hspace{5mm} \textsuperscript{2}University of Maryland
}

\begin{document}

\maketitle

\begin{abstract}
Speech perception involves storing and integrating sequentially presented items.
Recent work in cognitive neuroscience has identified temporal and contextual characteristics in humans' neural encoding of speech that may facilitate this temporal processing.
In this study, we simulated similar analyses with representations extracted from a computational model that was trained on unlabelled speech with the learning objective of predicting upcoming acoustics. 
Our simulations revealed temporal dynamics similar to those in brain signals, implying that these properties can arise without linguistic knowledge. 
Another property shared between brains and the model is that the encoding patterns of phonemes support some degree of cross-context generalization. 
However, we found evidence that the effectiveness of these generalizations depends on the specific contexts, which suggests that this analysis alone is insufficient to support the presence of context-invariant encoding.

\textbf{Keywords:} 
speech processing; speech representations; computational model
\end{abstract}

\section{Introduction}
Many perceptual processes involve tracking items that occur sequentially and integrating them to extract information.
One such process is speech perception, where successive phones\footnote{We use ``\emph{phones}" to refer to audio segments that form the basic units of speech, which are instantiations of the abstract ``\emph{phoneme}" categories. }, or speech sounds, are stored and combined before they are mapped onto lexical items. 
Although this process is typically effortless for human listeners, it is a non-trivial task because neighboring phones blend into each other due to co-articulation, and the same set of phonemes can form multiple words (e.g. \emph{cats, task}). 
There has been considerable recent interest in studying the neural representations that support this process \cite{gwilliams2022neural,khalighinejad_dynamic_2017,mesgarani_phonetic_2014,yi_encoding_2019,hamilton_revolution_2020}. 

Here we use computational modeling to better understand the representations that support listeners' temporal processing of speech.  
Our focus is on a recent study by \citeA{gwilliams2022neural} that investigated this question by analyzing MEG recordings from human listeners.  
\Getal\ replicated previous findings that the brain processes multiple phones simultaneously, showing that brains simultaneously encode at least three consecutive phones.  
They further showed that the encoding patterns of each phone are not static, but rather evolve over time. 
Additionally, they explored the extent to which each phone is encoded independently to its neighboring phones, and concluded that at least some part of the phone encoding is context-invariant.
These characteristics of neural speech processing are likely to play a role in supporting human listeners' ability to integrate information across time, but it is not known how or why these characteristics arise.

Our simulations build on \Getal's findings by performing similar analyses on representations extracted from a self-supervised computational model that is trained to predict the upcoming acoustics based on context.  
We find that, like humans, the model processes multiple phones simultaneously and its representations of those phones evolve over time. 
Moreover, like humans, both our model and an acoustic baseline show at least some context invariance.  
This suggests that many of the properties of speech representations that \Getal\ found can arise through predictive learning, without requiring prior linguistic knowledge.  
We further identify properties in our models' representations that appear to deviate from those found in humans, and thus may not be direct representational consequences of the prediction task.  
As a whole, our work illustrates how modern architectures from speech technology can help provide insight into the factors that shape speech representations in human listeners.

\section{Method}
This work consists of three simulations that tested (1) the window of phonetic decodability (2) the time course of phone encoding and (3) cross-context generalization of phonetic decoders. 
Fundamental to all three simulations is training decoders to track phonetic encodings in representations from a predictive speech model. 
In this section, we first introduce the model and the corpus we used to extract representations and acoustic features to be examined in our simulations. 
Then we describe how we decoded phonetic categories from the representations.
The analysis procedures and the results are grouped by simulation and presented following this section.

\subsection{Model and representations}

In this work, we used a recurrent neural network model that was pre-trained with self-supervised learning (SSL), meaning that it learns just by being exposed to raw input data (in this case speech audio), with no external training signal or annotated labels. 
SSL models have become widespread as representation learning methods in machine learning of speech and language \cite{devlin_bert_2019,baevski_wav2vec_2020,hsu_hubert_2021}, and have also been evaluated as models of speech perception in recent years \cite{millet_self-supervised_2022, millet_toward_2022,tuckute2023many}.

The SSL model we used is based on the contrastive predictive coding (CPC) framework \cite{oord2018representation}, and uses cognitively plausible mechanisms (prediction and error-driven learning) to learn a 512-dimensional vector representation for each 10ms frame (time slice) of the input. The model's learning objective is to find representations that can be effectively used to predict the representations for upcoming frames (specifically, the next 12 frames, or 120ms of speech). 

The model was implemented and trained by \citeA{nguyen_zero_2020}, and consists of 3 LSTM layers on top of 5 convolutional layers, all trained jointly.  
The model was trained on 6000 hours of audiobooks from the ``clean-light" subset of the Librilight corpus \cite{kahn2020libri}.

In our simulations, we applied our analyses to CPC representations and acoustic features extracted from a distinct set of audiobooks, the ``dev-clean" subset of Librispeech \cite{panayotov_librispeech_2015}.
The subset contains 8 minutes of read speech from 40 speakers, of which there are 21 females and 19 males.
We obtained time-aligned phoneme labels of the audio through forced alignment with the transcriptions. 
39 phonemes occurred in the transcriptions, of which there are 15 vowels and 24 consonants. 

We extracted the output of the second LSTM layer as the model's representations, since that layer gave the best performance in phone classification \cite{nguyen_zero_2020}.
For acoustic features, 40-dimensional logmel spectrogram features were extracted using the torchaudio package\footnote{\texttt{torchaudio.compliance.kaldi.fbank}}.

\subsection{Decoding for phonetic information}
To analyze the representations learned by the CPC model, we fit ridge regression models to identify the phoneme label of each representation vector, and computed the accuracy of these decoders on a held-out test set.
The choice of training and testing data varies by simulation, according to the specific question under investigation.
We consider phonetic information to be represented if the decoding accuracy is higher than the majority class baseline (most common phoneme label).
The decoders are implemented using the ridge regression function from the sklearn package with the default regularisation parameter.
Phone boundaries and labels are obtained through forced alignment with an acoustic model created according to the official Kaldi recipe for LibriSpeech data\footnote{\url{https://github.com/kaldi-asr/kaldi/blob/master/egs/librispeech/s5/run.sh}}.

\Getal\ pointed out that phonetic information is partly confounded with low-level acoustic properties such as amplitude and pitch, and they therefore preprocessed their neural recordings using a linear model to regress out these two factors.
For CPC representations, we performed preprocessing by training two ridge regression models to predict amplitude and pitch values from each representation, and regressing out the variance in the direction given by the regressors' coefficients. 
While this operation made little difference to the results of our experiments, all the results reported in this paper were obtained with preprocessed CPC representations. 
We followed \Getal\ in not preprocessing logmel features.

\section{The window of phonetic decodability}
In this simulation, we examine the time window during which the phoneme category of a phone can be decoded from model representations or acoustic features.
While the average duration of a phone is around 80ms, phoneme identity could be decodable for longer than 80ms due to coarticulation, although a decodable window considerably longer would imply that information about multiple phones is maintained at the same time. 
Neuroimaging studies have found that phonetic features and/or phone identity are decodable from brain recordings for between 200-400ms, starting around 10-50ms after phone onset \cite{khalighinejad_dynamic_2017,gwilliams2022neural}.

\subsection{Procedure}
For each phone token, we considered a time window of 1600ms centered at phone onset, which corresponds to 160 10ms frames. 
A separate decoder is trained for each of the 160 time steps to determine whether phones are decodable up to 800ms before or 800ms after their onset.
For example, the -70ms decoder is trained on all frames occurring 70ms prior to a phone boundary, and must predict the phoneme label of the phone that starts 70ms later.
We used 5 utterances from each of the 40 speakers to train the decoders and 5 more for testing. 

\subsection{Results and Discussion}
\begin{figure}[h]
    \centering
    \includegraphics[width=0.9\columnwidth]{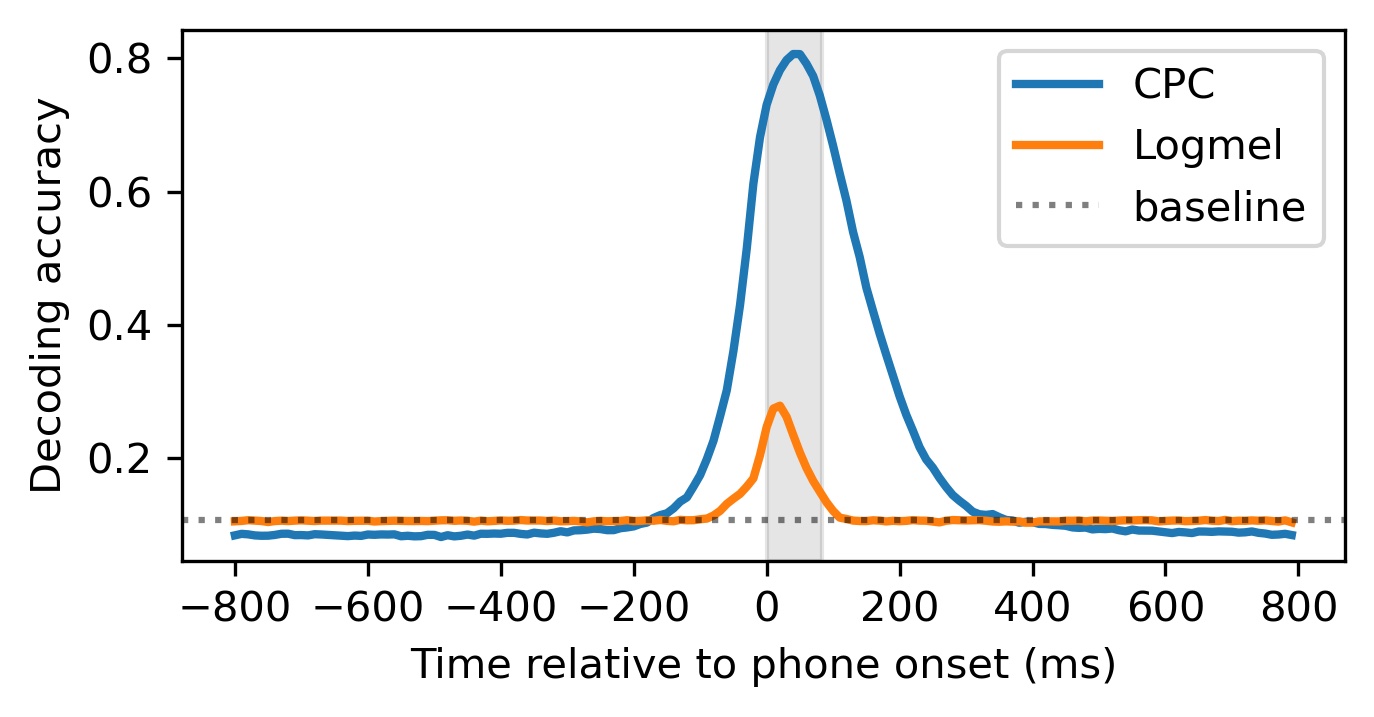}
    \vspace{-0.2cm}
    \caption{Accuracy of decoding for phoneme categories with CPC representations and logmel features. The shaded area represents the average duration of a phone.}
    \vspace{-0.4cm}
    \label{fig: decoding window}
\end{figure}

As shown in Figure \ref{fig: decoding window}, phones started to be decodable from CPC representations 180ms before phone onset and remained decodable for 540ms.
Thus, like human brains, the model maintains phone representations for far longer than their duration, implying that multiple phones are represented simultaneously (as explored further in the next simulation). 
In contrast, decoders trained on logmel features achieved accuracy above baseline 110ms before phone onset and dropped to baseline 230ms later.
While the window of decodability from acoustic features is also longer than the average phone duration of 84ms, it is still far shorter than that of the CPC representations.
We attribute the long window from acoustic features to a combination of coarticulation effects, variable duration of phones (i.e., the decoder will have a longer decodability window for phones that are longer than average), and perhaps some errors in the forced alignment. In future work we plan to replicate our experiment on hand-aligned data and consider durations of phones in our analysis.

Two other points are worth noting regarding these results.
First, phones can be decoded with a much higher accuracy from CPC representations than from acoustic features, which implies that different phones become more linearly separable in the learned  representations. 
Second, CPC representations seem to support predictive decoding, since phones are decodable 70ms prior to logmel features.
While this is not surprising given that the model is trained to predict the upcoming speech, it does differ from the findings of neuroimaging studies---a point we return to in the General Discussion.

\section{The time course of phone encoding}
Having found that multiple phones are encoded simultaneously in CPC representations, the question we aim to answer in this simulation is: \emph{how} does the model maintain information about successive phones without interference between them?

\subsection{Procedure}
One way of exploring this question with phonetic decoders is through \emph{temporal generalization} (TG) analysis.
After training decoders for each time step as in the previous analysis, we apply each decoder to all the time steps to test generalization.
Since each decoder has learned the most informative neural patterns for identifying the phonetic category at a certain moment, the extent that each decoder generalizes to neighboring time steps would reflect the time course of the encoding patterns.
The result of TG analysis is a training $\times$ testing matrix, which we visualize as a contour plot.  

To enable closer comparison with \Getal's results, we followed them in splitting the phone tokens according to their position in the word and performing TG analysis for each set.
This resulted in four TG matrices corresponding to the first to the fourth phone within each word (denoted p1--p4), which were visualized in the same plot (Figure \ref{fig: temporal generalization}), with matrices 2--4 shifted to the right by the average duration of their preceding phones.
Since the baseline decoder accuracy is 0.11, we plot contours at accuracy levels of 0.2---a clear improvement over the baseline but still far below maximum decoder accuracy (when training and testing time are the same)---and 0.4.

Since this is a computationally intensive analysis, we only used representations extracted from speech produced by two female and two male speakers, which are 32 minutes in total .

\subsection{Results and Discussion}
\begin{figure*}[!tb]
    \centering
    \includegraphics[width=2\columnwidth]{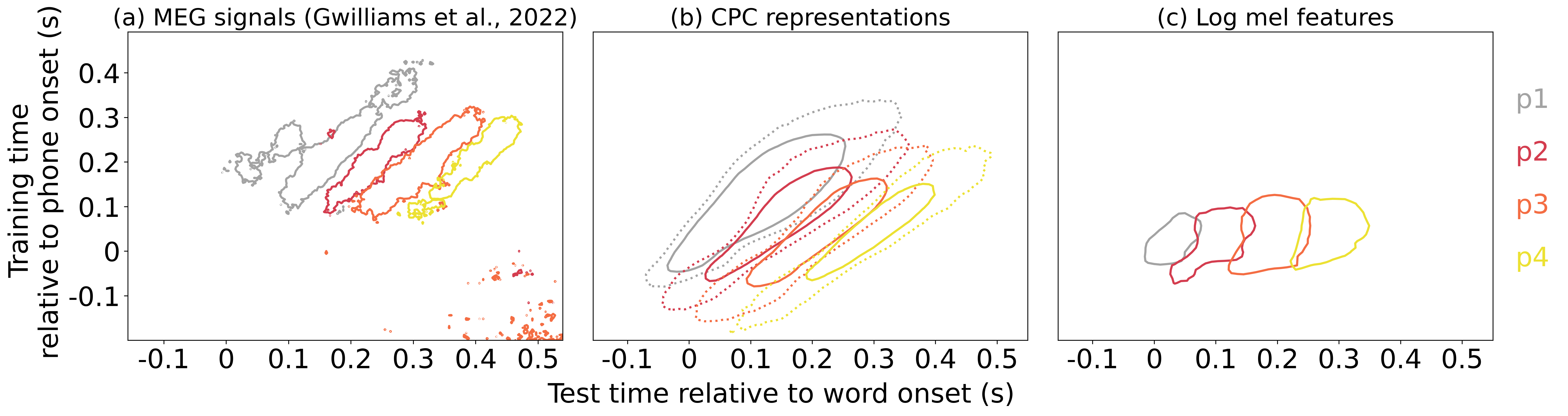}

   \caption{Temporal generalization (TG) results superimposed for 4 phone positions, the first to the fourth phone in each word (p1-p4), obtained with (a) MEG signals with contours at \emph{t}-value = 4 (reproduced from source data provided by \Getal); (b) CPC representations, with accuracy contours at 0.4 (solid) and 0.2 (dotted); and (c) Log mel features, with accuracy contours at 0.2.}
    
    \label{fig: TGours}
\label{fig: temporal generalization}
\vspace{-0.4cm}
\end{figure*}

As shown in Figure \ref{fig: temporal generalization}b, the temporal generalization patterns of CPC representations resemble those of the MEG signals (replotted in Figure~\ref{fig: temporal generalization}a) in two important aspects. 
Firstly, for all four positions and regardless of accuracy threshold, the diagonal axis of the contour is much longer than any of its horizontal widths.
While the diagonal axis signifies the period that a phone is decodable from the representations, the horizontal slices of the contour represent the duration that each neural pattern persists.
For example, with an accuracy threshold of $0.4$, a phone occurring in p1 to p4 is decodable for more than 200ms on average, while each specific neural pattern is maintained for no longer than 100ms.
In other words, the encoding pattern of a phone evolves dynamically throughout the period that it is decodable.

The second similarity between TG patterns of CPC and brains is that word-initial phones remain decodable for a longer period than phones in later positions.
Granted that p3 and p4 had fewer training samples than p1, which could be why p3 and p4 had narrower decodable windows.
However, p1 and p2 had almost the same number of training samples (p1:4793, p2:4757, p3:3202, p4:1852), and the distribution of phonetic categories at p2 even had lower entropy than p1 (p1:4.45 bits, p2:4.24 bits), although p1 is just slightly longer in average duration (p1:83.5ms, p2:81.9ms, p3:84.3ms, p4:83.2ms).
Still, the difference in the duration of decodability between p1 and p2 could imply that word-initial phones are maintained longer in the CPC model.

Neither of these properties was observed in the TG matrices of logmel features (Figure~\ref{fig: TGours} (c)), which means that they were acquired by the CPC model during self-supervised training.

Figure~\ref{fig: TGours}(b) also shows an interesting effect related to the model representation of upcoming phones. 
Note that contours dropping below 0 on the y-axis indicate decoders that can successfully predict the upcoming phone after being trained on representations prior to the phone onset. 
While this effect was already noted in our first simulation, the TG plots reveal an additional subtlety, which is that phones in later word positions seem to be predictable further in advance (contours are lower on the y-axis). 
This effect aligns with the long-standing observation that transitions between sub-word units are more predictable within words than at word boundaries, a fact that has been hypothesized to help infants begin to segment words \cite{saffran_statistical_1996}. 

\section{Generalizing decoders across contexts}
While the previous simulation addresses the interaction between consecutive phones in their temporal dynamics, the third simulation was targeted at \emph{how} neighboring phones are encoded.
Specifically, we aim to examine whether phones are encoded in a context-invariant manner in the representations.
We say there is context-invariant encoding if at least some part of the representations depends only on the current phoneme category and remains constant regardless of different surrounding contexts.
Since the acoustics of a phone are affected by co-articulation with neighboring phones, neighboring phones might be impossible to fully disentangle in the representations.
Alternatively, it might be that context-invariant patterns of each phone can be extracted through predictive learning.

We examined the presence of context-invariant phonetic encoding by testing cross-context generalization, namely training phonetic decoders for phones occurring in one specific context and testing them on phones in other contexts.
The generalization tests have three possible outcomes.
If the decoders fully generalize to all other contexts, it would be strong evidence for the notion that context-invariant encodings are present.
On the other hand, the decoders may fail to generalize at all, which would imply that phones are encoded as a whole with their co-articulatory neighbors. 
In between the two extremes is the possibility that there is some degree of generalization.
This situation is what \Getal\ found in neural representations, and concluded from it that the representations are at least partly context-invariant. 
However, we argue that this finding is more difficult to interpret, since partial generalization could simply be due to acoustic similarity between the phonetic realizations of the same phoneme in different contexts, with no additional context-invariant encoding being present. To begin to tease apart this question, we compare the level of contextual generalization achieved by learned representations against those of the acoustic features.

\subsection{Procedure}
We defined contexts in two ways: position in a word and phonetic context. 
Neural data are available only for the former, since this was the definition of context used by \Getal.
Choosing training and test cases for the latter can pose a challenge, since different contexts tend to feature distinct distribution of phones.
To evaluate generalization in a controlled setting, we chose to focus on decoding phoneme labels for vowels only, so that the class distribution is relatively consistent across different phonetic contexts as well as positions.

For cross-position generalization, we worked with four phone positions, p1--p4 as before. 
For the different phonetic contexts, we simplified the analysis by only considering the manner of articulation of the previous and the following phone, i.e., plosive, fricative, or nasal.
This yielded six contexts, each denoted as \emph{preceding phone\_\_following phone}, e.g., \emph{plosive\_\_plosive}. 
4500 phone samples were subsampled for each context type across all 40 speakers, with 80\% of those used for training decoders and the remaining 20\% for testing generalization.
We used the same number of subsamples and train-test split ratio for the four position classes.
Decoders were trained for each position/context class, and then tested on all of the position/context classes.

\subsection{Results and discussion}

\begin{figure*}[!tb]
\centering
\begin{subfigure}{.5\columnwidth}
\begin{tikzpicture}
    \node (img) at (0, 0) {\includegraphics[height=3cm]{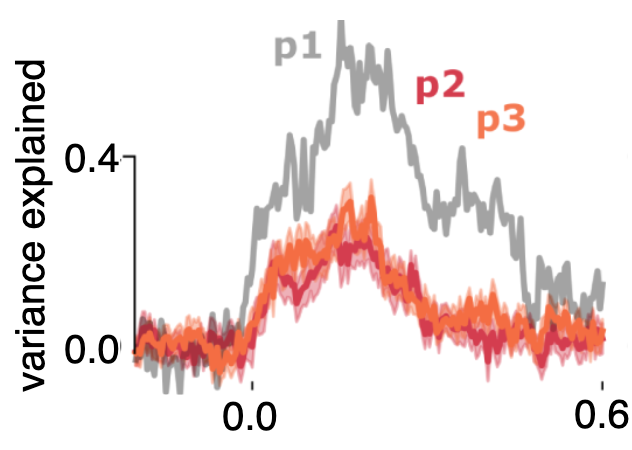}};
    \node at (0.5, -1.7) {\footnotesize Time relative to phone onset (s)};
\end{tikzpicture}
\caption{Results from \Getal}
\label{fig: Getal p1}
\end{subfigure}
\hspace{.3in}
\begin{subfigure}{1.2\columnwidth}
    \centering
\includegraphics[width=10cm]{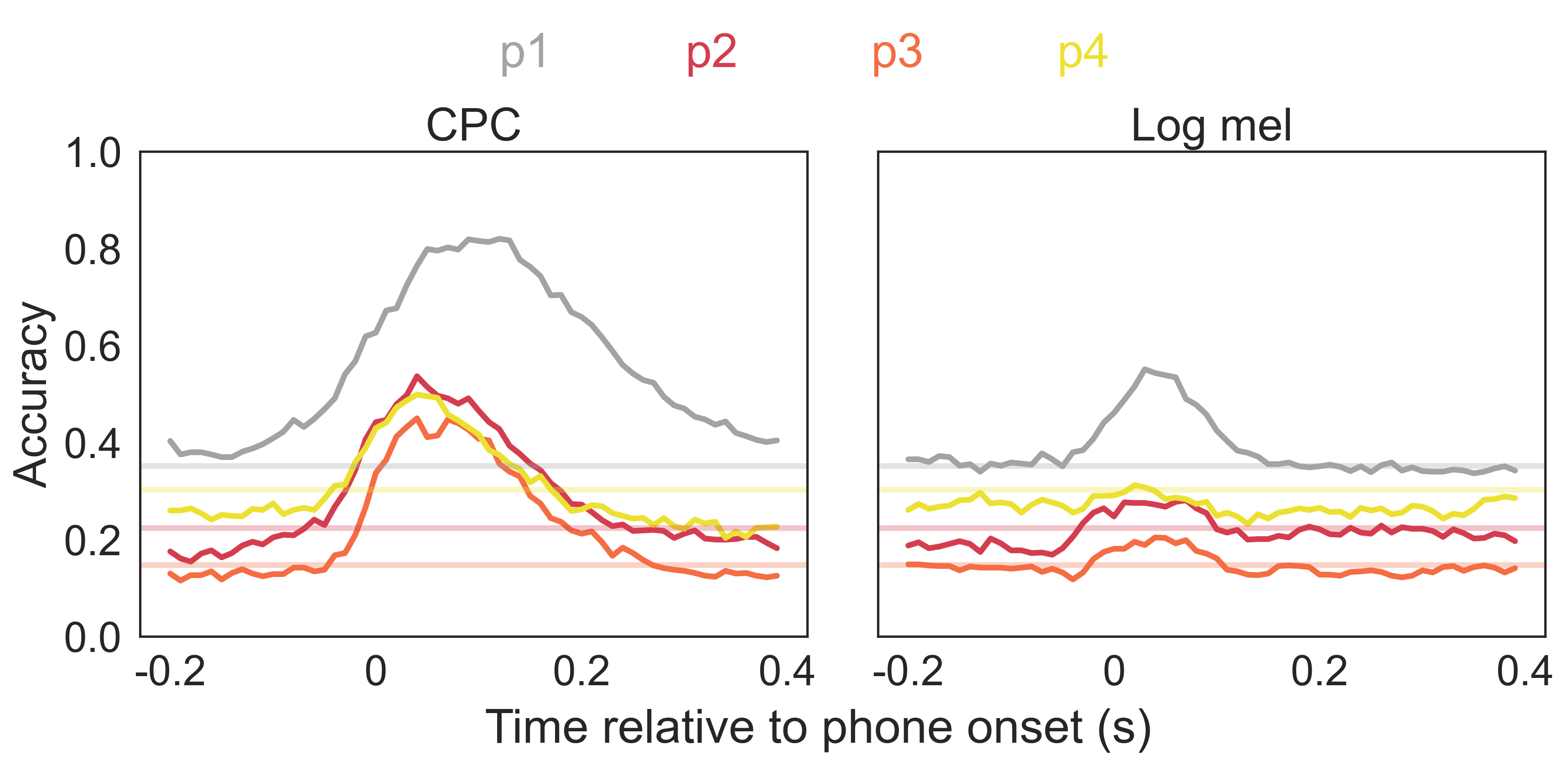}
\caption{Decoders from our simulations.}
    \label{fig: trained on p1}
\end{subfigure}
\caption{Generalizing from word-initial position to other word positions: (a) results from brain recordings (taken
from \Getal\ 2021) and (b) accuracy of our decoders. Decoders are trained on word-initial vowels (p1) and tested on vowels in p1-p4. The lefthand plot shows decoding accuracies for model representations, and the righthand plot for acoustic features.  The faded lines are the baseline accuracy for each position obtained by picking the most common vowel category in the training set. (Note that /a/ is the most common category in all positions, but to differing degrees, which leads to the different baselines.)}
\vspace{-0.4cm}
\end{figure*}

We first look at the part of our analysis that is analogous to the experiment in \Getal, namely training decoders on word-initial phones and testing on other positions.
From Figure \ref{fig: trained on p1} (left), we can see that the decoders trained on p1 showed significant generalization effects on p2, p3, and p4---that is, the curves for each of those positions are well above their respective baselines, which are  
computed as the accuracy obtained by picking the most common vowel in the training data (in this case, at p1) when evaluating on p1--p4. 

\begin{figure}[!h]
    \centering
\begin{subfigure}{\columnwidth}
    \centering
\includegraphics[width=\textwidth]{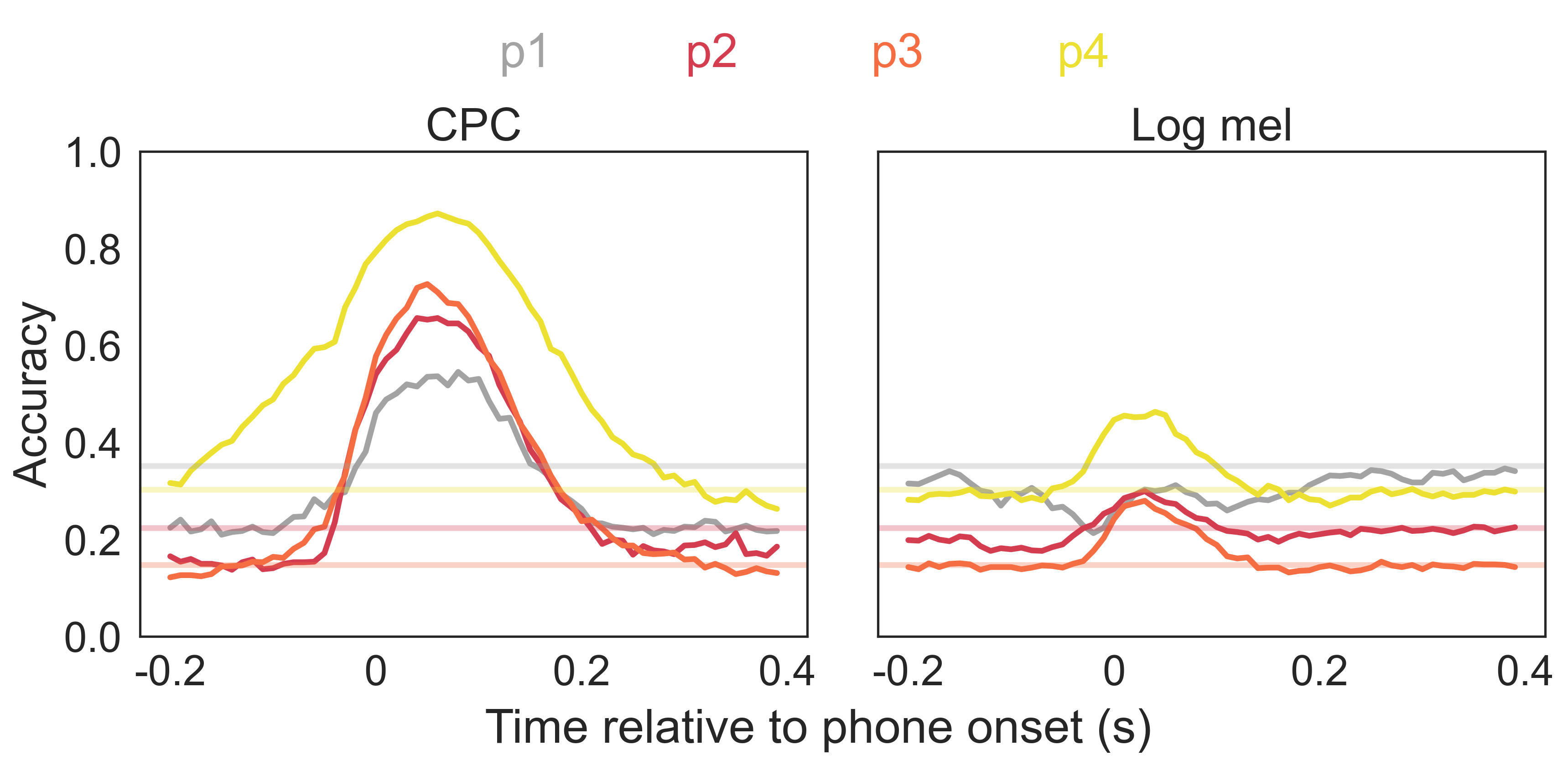}
\caption{Decoders trained on vowels in word position 4 (p4) and tested on other vowel positions. }
 \label{fig: trained on p4}
\end{subfigure}
\begin{subfigure}{\columnwidth}
    \centering
\includegraphics[width=\columnwidth]{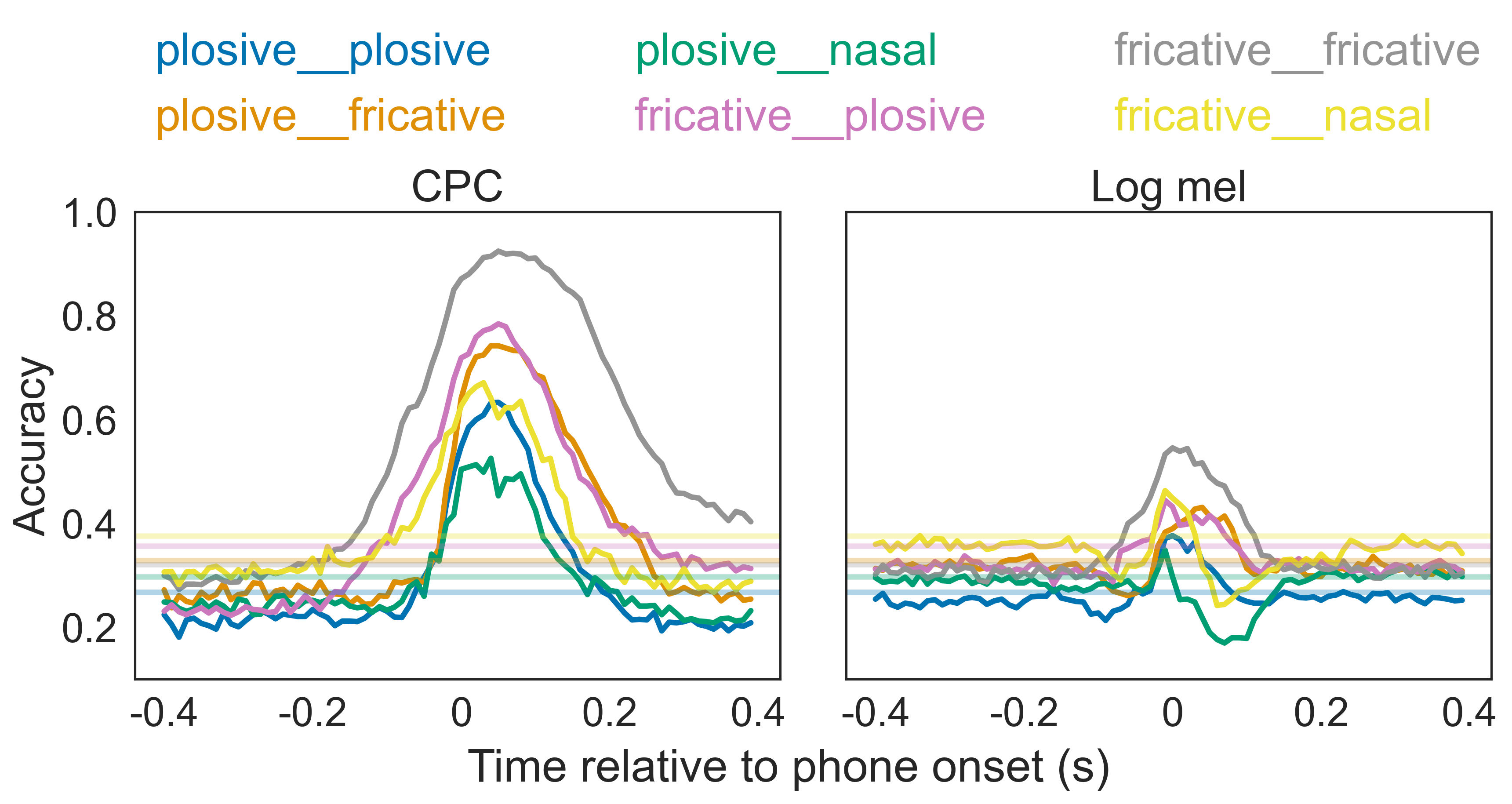}
\caption{Decoders trained on vowels surrounded by two fricative sounds and tested on vowels in five other contexts. }
\label{fig: trained on ff}
\end{subfigure}
\caption{Additional generalization tests across (a) different phone position and (b) different phonetic contexts. The faded lines represent the baseline accuracy obtained by picking the most common vowel at the training position/context.}
\vspace{-0.4cm}
\end{figure}

This result is qualitatively similar to \Getal's results (Figure \ref{fig: Getal p1}). However, as noted above, to better interpret this level of generalization, we also need to consider the extent to which acoustic similarity alone might explain patterns of generalization in the model representations. First, we note that for the decoder trained on acoustic features from p1, there is also some (albeit minimal) degree of generalization to p2--p4, as shown in Figure~\ref{fig: trained on p1} (right). In addition, Figure~\ref{fig: trained on p4} illustrates how the degree of generalization depends on both the training and testing sets: for example, decoders trained on p4 and tested on the other three positions generalize somewhat better than those trained at p1.

Turning to generalization across phonetic contexts,
we can again see from Figure \ref{fig: trained on ff} (left) that the decoders trained on \emph{fricative\_\_fricative} generalized to different degrees when tested on other context types. 
Specifically, the improvement over baseline in the test context of \emph{plosive\_\_nasal} was relatively modest.
Meanwhile, Figure \ref{fig: trained on ff} (right) shows that the decoders trained for logmel features also struggled to generalize to the same test context. 
A possible explanation is that the acoustics of vowels in this context differed more significantly due to nasalization.
This hypothesis is consistent with the fact that the generalization for logmel features in the test context \emph{fricative\_\_nasal} was initially strong around phone onset before the performance dropped below baseline, since this test context shared the same preceding phone as the training context but had a nasal as the following phone.
This pattern of dropping below and then returning to the baseline was not observed in the results for CPC representations.
Nevertheless, the results considered so far suggest that the generalization effects in CPC representations may be dependent on acoustic similarities in general. 

\begin{figure}
\centering
\includegraphics[width=\columnwidth]
{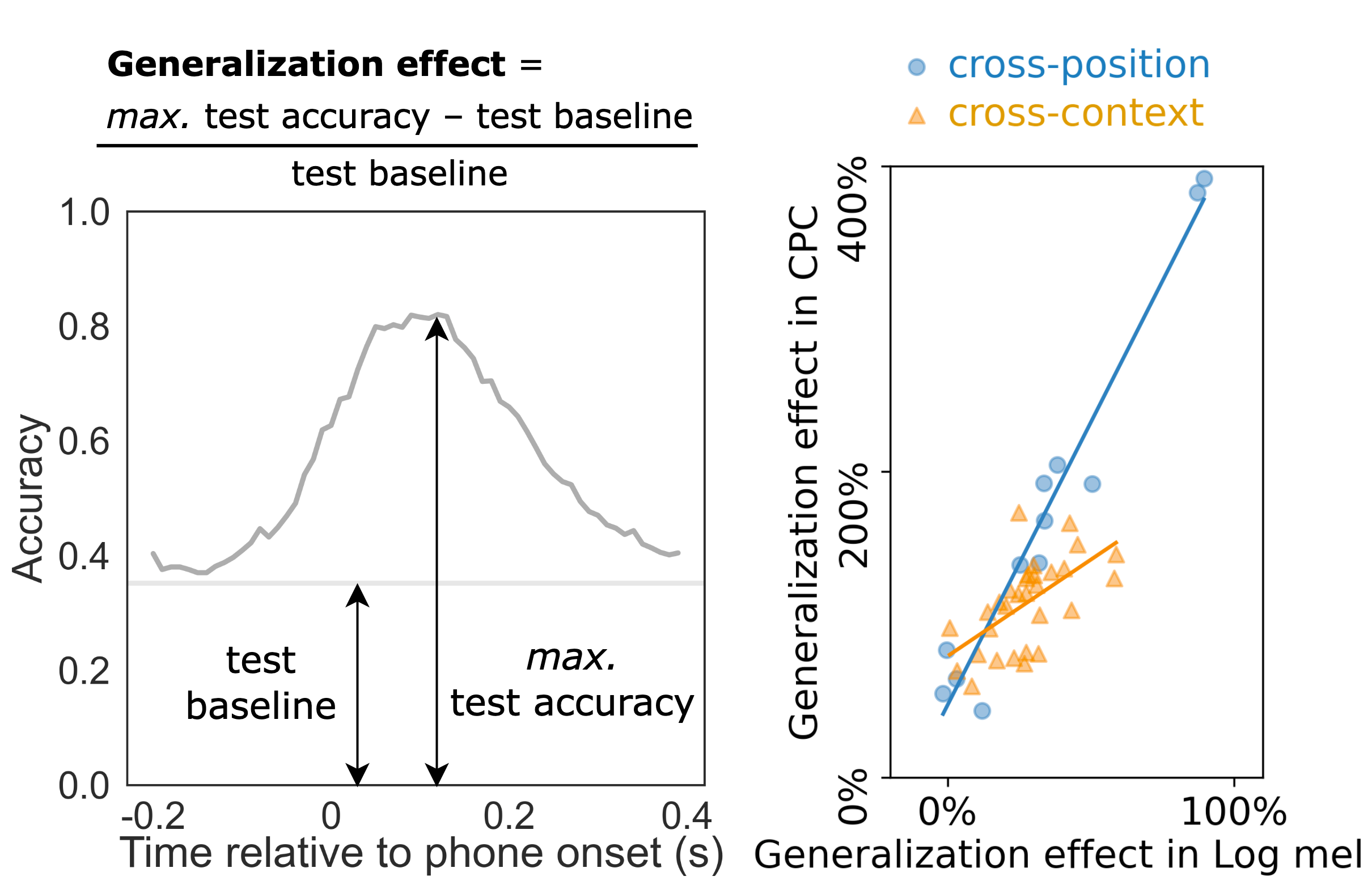}
\caption{Cross-position and cross-context generalization effects in log mel features and CPC representations correlate positively. Each circle represents the generalization effects on a test position/context. }
\label{fig: correlation}
\vspace{-0.4cm}
\end{figure}

To evaluate this hypothesis more systematically, we quantified the degree of generalization in CPC or logmel for each (training, test) context pair, as illustrated in  Figure~\ref{fig: correlation}(left).
The relationship between generalization effects in CPC and in logmel was then visualized in Figure~\ref{fig: correlation}(right), where each data point corresponds to a particular (training, test) context pair, with the (x,y) coordinates indicating that pair's generalization effect in CPC and logmel respectively. We only include pairs where the training and test sets are different contexts.

The 12 blue dots in Figure \ref{fig: correlation} show the generalization effects across phone positions, which has a Pearson correlation of 0.97 ($p<10^{-7}$).
The 30 orange data points represent generalization effects across phonetic contexts, with Pearson correlation of 0.60 ($p<10^{-3}$).
Both correlations are strong enough to suggest that both cross-position and cross-context generalization in CPC representations depend on similarities between the acoustics of the training and the test contexts. 
While it is still possible that the learning induces some additional context invariance beyond acoustic similarity, the partial generalization found here does not seem sufficient to conclude that.

\section{General Discussion}
This paper analyzed the representations in a computational model to gain insight into the neural representations that support human listeners' encoding of phonetic information across time. 
We found that a self-supervised predictive model simulated key aspects of phonetic encoding found in brain signals, specifically with respect to temporal dynamics and contextual invariance. 
In particular, we showed that, like the neural representations but unlike the acoustic signal, the model representations support tracking of multiple successive phones simultaneously using a rapidly evolving representation. 
Furthermore, like the neural representations, the model's representations of phones can generalize across contexts to some extent. 
However, unlike previous work, we also examined cross-context generalization in decoders trained on acoustic features, and found that these, too, show partial generalization. 
Moreover, the extent of cross-context generalization in the model representations is strongly correlated with acoustic similarity (i.e., generalization using acoustic features). 
We conclude that there is little evidence that the model has learned context-invariant representations beyond acoustic similarity. 
This result also suggests that further analysis of the neural data is required before concluding that context-invariant representations have been found. 
These concerns aside, the success of this model in simulating key aspects of the brain data suggests that a purely predictive model with no top-down supervision could potentially explain some of the computational principles underlying the processing of speech and other sequential processing tasks.

As well as the above similarities between model and brain representations, we also found one important difference between the temporal dynamics of our model and that found by brain imaging studies to date: our model's representations encode predictions about the upcoming phone, whereas neuroimaging studies of continuous speech have mainly found evidence of phonetic decodability only after phone onset \cite{mesgarani_phonetic_2014,khalighinejad_dynamic_2017,gwilliams2022neural}. A notable exception is that word-initial phones, but not those in other positions, appear to be decodable at phone onset in \Getal's study.
Given the predictive nature of our model, it is not surprising to find predictive representations, but our finding does highlight the surprising {\em lack} of such findings in brain imaging studies. It is well-established behaviorally that human listeners make predictions about upcoming linguistic material \cite{ryskin_prediction_2023}. Recent studies have also argued that fMRI and ECoG data from listening to speech and reading text are well-modelled by predictive models of text \cite{schrimpf_neural_2021,goldstein_shared_2022,caucheteux_evidence_2023}, and have found evidence that specific words are decodable from ECoG recordings prior to word onset, indicating word-level predictive representations \cite{goldstein_shared_2022}. An important question for future work is therefore whether current tools are simply not sensitive enough to identify lower-level predictive representations in the brain, or whether these are indeed absent---implying that prediction operates only over higher-level representations.

This research provides insight into the factors that shape speech representations in humans and machines, but leaves a number of open questions.  For example, the recurrent neural network architecture used here differs from the transformer architecture that is now more often used in speech technology, and future research can explore whether attention-based architectures, such as transformers, also yield speech representations that exhibit similar characteristics.  The field's interest in characterizing human speech representations also stems, in large part, from a desire to understand how the structure of these representations support higher-level tasks, such as word recognition.  Here, computational models can be particularly useful in enabling a controlled comparison between different representations, especially because representations from these models have enabled large advances in the accuracy with which speech technology performs those higher-level tasks.

Finally, although this work is primarily aimed at cognitive scientists, it may also be relevant for speech technology researchers, since the model we use shares many characteristics with current state-of-the-art SSL models for speech. 
Within the speech technology community, analyses of
DNN models' phonetic representations have mainly focused on how accessible phonetic or phonemic information about the current phone is, either in different models \cite{ma_probing_2021} or across model layers \cite{belinkov_analyzing_2017,chrupala_analyzing_2020,cormac_english_domain-informed_2022,martin_probing_2023,ten_bosch_phonemic_2023,pasad_layer-wise_2021,pasad_comparative_2023}.
There is less work on {\em how} such information is encoded, although a few researchers have used clustering or visualization to investigate this question \cite{nagamine_exploring_2015,nagamine_role_2016,de_seyssel_probing_2022,wells_phonetic_2022}, while others have shown that formants are represented in a structured way \cite{choi2022opening} and that phonetic and speaker information are represented in orthogonal subspaces \cite{liu_self-supervised_2023}.
We know of no analyses examining the temporal dynamics of speech representations, or of any that investigate context-invariance by testing decoders for generalization to unseen contexts. These questions and methods, inspired by work in cognitive neuroscience and demonstrated on SSL models by this study, could prove fruitful for other researchers interested in analyzing speech model representations for their own sake.

\vfill

\section{Acknowledgements}
This work was supported in part by the UKRI Centre for Doctoral Training in Natural Language Processing, funded by the UKRI (grant EP/S022481/1)
and the University of Edinburgh.
\bibliographystyle{apacite-mod}

\setlength{\bibleftmargin}{.125in}
\setlength{\bibindent}{-\bibleftmargin}

\bibliography{references,refs-sgwater-clean}

\end{document}